\documentclass[10pt,twocolumn,letterpaper]{article}

\usepackage{cvpr}              
\definecolor{cvprblue}{rgb}{0.21,0.49,0.74}
\usepackage[pagebackref,breaklinks,colorlinks,allcolors=cvprblue]{hyperref}
\usepackage{bm}
\usepackage{colortbl}
\usepackage{graphicx}
\usepackage{multirow}
\usepackage{xcolor}
\usepackage{footnote}
\usepackage{pifont}
\usepackage{algorithm}
\usepackage{algorithmic}
\usepackage[accsupp]{axessibility}

\usepackage{subcaption}
\newdimen\figrasterwd
\figrasterwd\textwidth

\def\paperID{} 
\def\confName{CVPR}
\def\confYear{2026}

\title{Self-Paced and Self-Corrective Masked Prediction for Movie Trailer Generation}

\author{
    \makebox[0.33\textwidth][c]{
        \begin{tabular}{c}
            Sidan Zhu \\
            {\small School of Computer Science and Technology} \\
            {\small Beijing Institute of Technology, Beijing} \\
            {\tt\small sidan\_zhu@bit.edu.cn}
        \end{tabular}
    }
    \makebox[0.33\textwidth][c]{
        \begin{tabular}{c}
            Hongteng Xu \\
            {\small Gaoling School of Artificial Intelligence} \\
            {\small Renmin University of China, Beijing} \\
            {\tt\small hongtengxu@ruc.edu.cn}
        \end{tabular}
    }
    \makebox[0.33\textwidth][c]{
        \begin{tabular}{c}
            Dixin Luo\thanks{Corresponding author.} \\
            {\small Beijing Institute of Technology, Beijing} \\
            {\small Key Laboratory of Artificial Intelligence} \\ 
            {\small Ministry of Education, Shanghai} \\
            {\tt\small dixin.luo@bit.edu.cn}
        \end{tabular}
    }
}

\begin{document}
\maketitle
\begin{abstract}
As a challenging video editing task, movie trailer generation involves selecting and reorganizing movie shots to create engaging trailers. 
Currently, most existing automatic trailer generation methods employ a ``selection-then-ranking" paradigm (i.e., first selecting key shots and then ranking them), which suffers from inevitable error propagation and limits the quality of the generated trailers. 
Beyond this paradigm, we propose a new self-paced and self-corrective masked prediction method called SSMP, which achieves state-of-the-art results in automatic trailer generation via bi-directional contextual modeling and progressive self-correction. 
In particular, SSMP trains a Transformer encoder that takes the movie shot sequences as prompts and generates corresponding trailer shot sequences accordingly. 
The model is trained via masked prediction, reconstructing each trailer shot sequence from its randomly masked counterpart. 
The mask ratio is self-paced, allowing the task difficulty to adapt to the model and thereby improving model performance.
When generating a movie trailer, the model fills the shot positions with high confidence at each step and re-masks the remaining positions for the next prediction, forming a progressive self-correction mechanism that is analogous to how human editors work. 
Both quantitative results and user studies demonstrate the superiority of SSMP.
Demo and code are available at: \url{https://github.com/Dixin-Lab/SSMP}.

\end{abstract}    
\section{Introduction}
\label{sec:intro}

Movie trailers play a vital role in the movie industry, serving as a key medium for attracting audiences and shaping their first impressions of a movie.
Typically, trailer editing involves iteratively selecting and arranging shot sequences to ensure coherence and harmony throughout the generated trailer, making it an exceptionally challenging and time-consuming task that requires semantic video understanding.
To make the entire process more efficient, numerous efforts~\cite{irie2010automatic, papalampidi2023finding, chen2004action, smeaton2006automatically, smith2017harnessing, argaw2024towards, praveen2025video, zhu2025weakly} have been made to automate the generation of movie trailers.

\begin{figure}[t]
    \centering
    \begin{subfigure}{0.9\linewidth}
    \includegraphics[width=\linewidth]{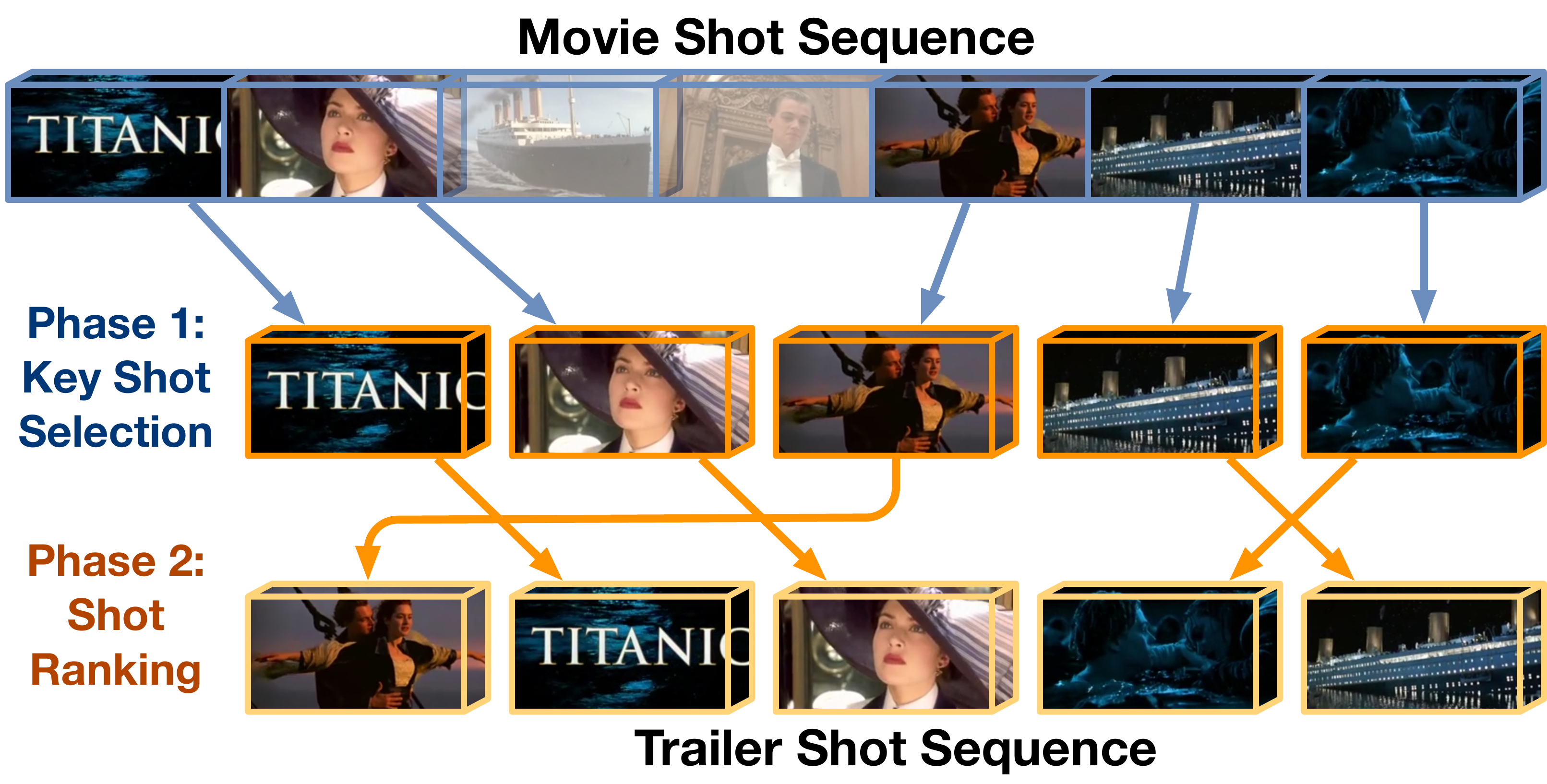}
    \caption{The selection-then-ranking paradigm}
    \label{fig:select_rank}
    \end{subfigure}
    \begin{subfigure}{0.9\linewidth}
    \includegraphics[width=\linewidth]{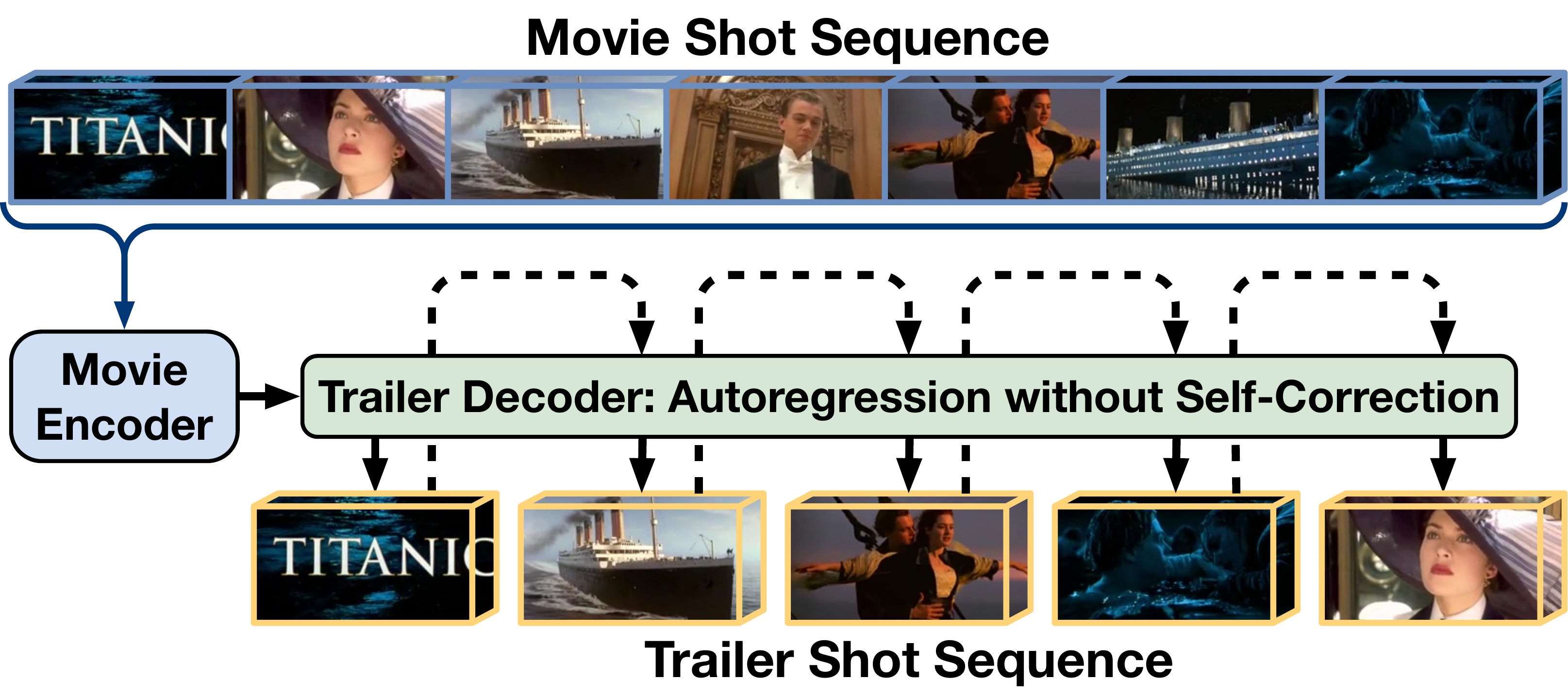}
    \caption{The recent auto-regressive paradigm}
    \label{fig:autogressive}
    \end{subfigure}
    \begin{subfigure}{0.9\linewidth}
    \includegraphics[width=\linewidth]{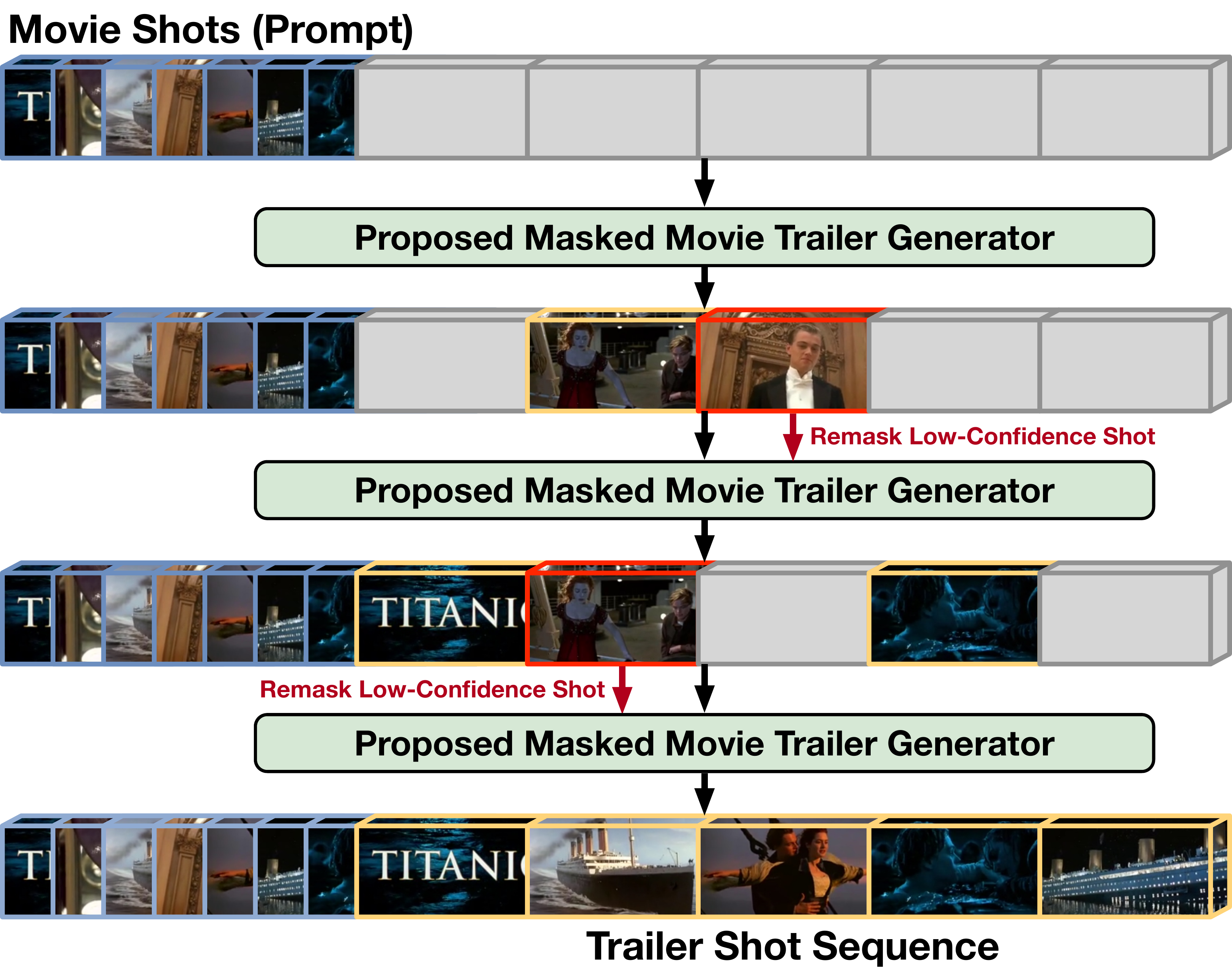}
    \caption{Our SSMP method}
    \label{fig:ours}
    \end{subfigure}
    \caption{Illustrations of different trailer generation paradigms. 
    Both classic selection-then-ranking and recent auto-regressive paradigms suffer from error propagation. 
    Our method mitigates this issue, modeling bi-directional context and achieving self-correction via progressive re-masking. 
    The shots used in this figure are from the movie \textit{Titanic}.}
\end{figure}

Early methods~\cite{irie2010automatic, papalampidi2023finding, chen2004action, smeaton2006automatically, smith2017harnessing, praveen2025video} are mostly rule-based, relying on additional manually-designed editing rules and thus suffering from poor generalizability.
With the development of deep learning, several data-driven methods have been proposed~\cite{wang2024inverse,argaw2024towards,zhu2025weakly,sun2025multi,jin2025tvmtrailer} to directly learn trailer composition patterns from paired movie–trailer data.
In principle, these methods yield the same ``\textit{selection-then-ranking}" paradigm shown in~\cref{fig:select_rank}, which decomposes trailer generation into two steps: shot selection and ranking.
This paradigm prevents the model from jointly reasoning about the semantic relevance and temporal continuity among shots. 
Recently, some attempts have been made to achieve selection and ranking jointly, formulating the task as a sequence-to-sequence problem~\cite{argaw2024towards, jin2025tvmtrailer}.
As shown in~\cref{fig:autogressive}, these methods predict trailer shots conditioned on the given movie in an auto-regressive manner.
However, both selection-then-ranking and auto-regression lack self-correction mechanisms to revise early shot selections/predictions, which fundamentally differ from how human editors work, as professional editors repeatedly refine shot connections during the trailer editing process.
As a result, these two paradigms suffer from inevitable error propagation, which limits the quality of the generated trailers.

To address the above issues, we formulate the trailer generation task as a masked prediction problem and propose a novel \textbf{S}elf-paced and \textbf{S}elf-corrective \textbf{M}asked \textbf{P}rediction (\textbf{SSMP}) method.
In particular, SSMP learns a Transformer encoder-based trailer generator with a self-correction mechanism in a self-paced learning framework. 
As shown in~\cref{fig:ours}, the model takes the given movie shots as the prompt and reconstructs the masked trailer shots iteratively.
At each step, the model predicts shots of all masked positions simultaneously and re-masks low-confidence shots for subsequent iterations.
This generation process effectively alleviates the error propagation issue --- it considers shot selection and ranking jointly and incorporates a self-correction mechanism that revises the generated shots.
When training our model, we are inspired by self-paced learning strategies~\cite{NIPS2010_e57c6b95,jiang2014self}, designing a self-paced mask ratio scheduler to improve training efficiency, stability, and model performance.
The scheduler dynamically adjusts the mask ratio of the shot sequences according to the capability of the model in the training process.



To the best of our knowledge, our work makes the first attempt to $i)$ build a movie trailer generator with bi-directional contextual modeling and progressive self-correction and $ii)$ develop a self-paced learning strategy for trailer generation models.
We compare our method with representative video summarization methods and state-of-the-art trailer generation methods in various datasets.
Both objective and subjective evaluations demonstrate the superiority of our method compared to previous works.

\begin{figure*}[t]
    \centering
    \begin{subfigure}[t]{0.48\textwidth}
        \centering
        \includegraphics[width=\textwidth]{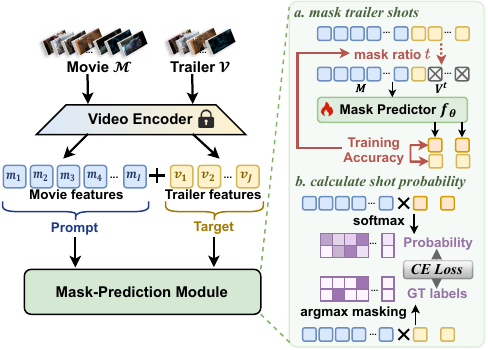}
        \caption{Training with a self-paced mask ratio scheduler}
        \label{fig:training}
    \end{subfigure}
    \hfill
    \begin{subfigure}[t]{0.48\textwidth}
        \centering
        \includegraphics[width=\textwidth]{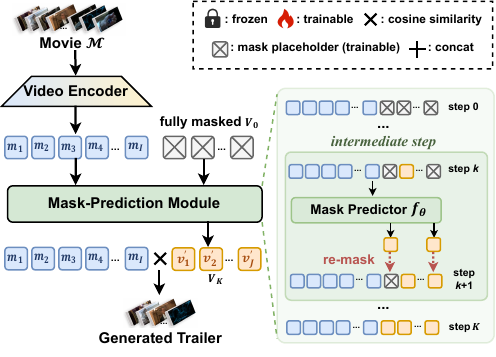}
        \caption{Generation with a self-correction mechanism}
        \label{fig:inference}
    \end{subfigure}
    \caption{An illustration of our SSMP method. (a) shows the training pipeline, including the self-paced masking and probability computation.
    (b) shows the generation pipeline, including iterative masked prediction and re-masking of partially masked positions.}
    \label{fig:scheme}
\end{figure*}
\section{Related Work}
\label{sec:relatedworks}


\subsection{Automatic Trailer Generation}
As aforementioned, early automatic trailer generation methods are rule-based, leveraging meta or side information to summarize common trailer grammar or editing templates.
These methods can extract key components (\eg, logo, theme music, and so on) of trailers from movies~\cite{irie2010automatic, kakimoto2018extraction}. 
Some of them further identify turning points and cinematic grammars from movies for guiding trailer generation~\cite{papalampidi2023finding, chen2004action}, which leads to useful editing rules for specific movie genres~\cite{smeaton2006automatically,smith2017harnessing, praveen2025video}.
Such rule-based methods suffer from poor generalization, since the predefined rules cannot cover various movie narratives and visual styles.

To address these issues, data-driven methods are proposed to learn trailer generators from movie-trailer pairs in supervised~\cite{rehusevych2019movie2trailer} or unsupervised~\cite{liu2015semi,wang2020learning} manner.
These methods commonly apply the ``selection-then-ranking" paradigm, selecting candidate shots for trailers and arrange them accordingly. 
Among them, various measurements are used as shot selection criteria, including the visual attractiveness~\cite{xu2015trailer}, the shot expressiveness~\cite{shaukat2024genvis}, the shot emotional intensity~\cite{hu2024decoupled}, and the shot relevance based on side information (\eg, viewer preference~\cite{lee2025hippo}, subtitles~\cite{hesham2018smart}, movie 
summaries~\cite{gaikwad2021plots, balestri2024trailer, balestri2024automatic,kawai2007automated}).
For shot ranking, most existing methods rely on external guidance~\cite{li2025shot}, such as music-video semantic consistency~\cite{sun2025multi,wang2024inverse,JCST-2212-13064, zhu2025weakly} and narrative assistance from large language models~\cite{balestri2024trailer, balestri2024automatic,li2025shots}.
However, the selection-then-ranking paradigm suffers from inherent limitations, as it decouples shot selection and ranking despite their strong interdependence.
To unify these two steps, the work in~\cite{argaw2024towards} draws inspiration from machine translation and formulates the task as a sequence-to-sequence prediction.
Building upon this work, the subsequent study~\cite{jin2025tvmtrailer} incorporates the movie’s main plot as conditional inputs, and further generates an accompanying soundtrack based on the generated trailer.
Similar to existing language models, these two methods apply an auto-regressive (AR) paradigm, predicting each shot based on the given movie shots and the previous predictions.
However, without any self-correction mechanism, these AR-based methods are still different from how human editors work --- human editors usually polish a trailer iteratively, repeatedly selecting and replacing shots across different shot positions instead of following a strictly sequential order.

\subsection{Masked Prediction for Generation}
As a powerful learning paradigm, masked prediction has been widely used in pre-training foundation models.
Typically, BERT~\cite{devlin2019bert} achieves a pre-trained large-scale language model by masked prediction, obtaining remarkable success across various natural language processing (NLP) tasks.
Motivated by BERT, CMLM~\cite{ghazvininejad2019mask} achieves conditional masked language modeling, and LLaDA~\cite{nie2025large} proposes a discrete diffusion model with a random mask ratio, allowing for contextually aware text generation.
The success in NLP motivates studies to explore masked prediction for image representation learning, which is often implemented by reconstructing masked pixels~\cite{he2022masked} and patches~\cite{bao2021beit,dong2023peco,chang2022maskgit}.
For video representation learning, the BEVT in~\cite{wang2022bevt} and the joint visual-linguistic models in~\cite {sun2019videobert,yang2020bert} are learned to reconstruct masked video tokens, both demonstrating the effectiveness of masked prediction. 
However, the current methods seldom investigate the impact of mask ratio on the model performance, and none of them consider constructing trailer generators via masked prediction. 
Our work fills this blank, learning a strong movie trailer generator via a self-paced masked prediction method.

\section{Proposed Method}
\label{sec:method}


In this study, we denote a movie and its corresponding trailer as $\mathcal{M}$ and $\mathcal{V}$, respectively.
Referring to previous methods~\cite{argaw2024towards, zhu2025weakly}, the movie and its trailer are segmented by the shot-boundary detector TransNet-v2~\cite{soucek2024transnet}, leading to two shot sequences, i.e., $\mathcal{M}=\{m_i\}_{i=1}^{I}$ and $\mathcal{V}=\{v_j\}_{j=1}^{J}$.
Typically, each trailer shot is sampled from the movie shots, i.e., $\mathcal{V}\subset\mathcal{M}$.
In this work, we formulate the movie trailer generation task as a masked prediction problem conditioned on the given movie.
As illustrated in~\cref{fig:scheme}, our model takes the movie shot sequence $\mathcal{M}$ and trailer shot sequence $\mathcal{V}$ as inputs, in which each shot is treated as a token.
The model captures both the underlying shot selection mechanism from the source movie and the sequential dependencies among the trailer shots through the reconstruction of masked trailer shots. 
When generating a trailer, the well-trained model takes the given movie shot sequence as a prompt and progressively fills the trailer sequence from all mask placeholders.
During this process, a self-correction mechanism is implemented as re-masking low-confidence shots iteratively.



\subsection{A Masked Prediction Framework for Training}
Inspired by BERT~\cite{devlin2019bert} and LLaDA~\cite{nie2025large}, our model is trained under a conditional masked prediction framework, where partial input tokens are masked through a forward process and reconstructed through a reverse process.
As shown in~\cref{fig:training}, we utilize a pretrained model ImageBind~\cite{girdhar2023imagebind} as our video encoder, extracting movie shot features $\bm{M}=[\bm{m}_i]\in \mathbb{R}^{I\times D}$ (\ie, the prompt) and trailer shot features $\bm{V}=[\bm{v}_j]\in \mathbb{R}^{J\times D}$ (\ie, the target), respectively, where $D$ denotes the feature dimension.
Each trailer shot is considered to originate from the movie shot that has the highest similarity score.
The corresponding ground truth labels are formulated as a matrix $\bm{G}=[g_{j,i}] \in \{0,1\}^{J\times I}$, \ie, for $j=1,...,J$ and $i=1,...,I$, we have $s_{j,i}=\frac{\langle\bm{v}_j, \bm{m}_i\rangle}{\|\bm{v}_j|\|\bm{m}_i\|}$ and  
\begin{eqnarray}\label{eq:indices}
\begin{aligned}
g_{j,\ell} =
\begin{cases}
1, & \text{if }\ell=\arg\sideset{}{_{i \in \{1, \dots, I\}}}\max s_{j,i}, \\
0, & \text{otherwise,}
\end{cases}
\end{aligned}
\end{eqnarray}
where $\langle\cdot, \cdot\rangle$ denotes the inner product operation.

We concatenate $\bm{M}$ and $\bm{V}$ into a single sequence and feed the sequence into the proposed mask predictor $f_{\theta}$, which is implemented as a Transformer encoder~\cite{vaswani2017attention}. 
Within this predictor, the prompt part is kept unchanged, while the target features are independently masked at the same ratio, resulting in a partially masked sequence $\bm{V}^t=[\bm{v}_j^t]\in \mathbb{R}^{J\times D}$, \ie, for $j=1,...,J$,
\begin{eqnarray}\label{eq:logits}
\epsilon\sim \text{Unif}[0,1],\quad
\begin{aligned}
    \bm{v}^t_j =
\begin{cases}
\text{MP}, & \text{if } \epsilon <t, \\
\bm{v}_j, & \text{otherwise},
\end{cases}
\end{aligned}
\end{eqnarray}
where $\text{MP}\in \mathbb{R}^{D}$ is the learnable mask placeholder, $t\in[0,1]$ is the mask ratio.
Accordingly, we denote the indices of the masked positions as $\mathcal{J}'\subset\{1,...,J\}$.
Taking $[\bm{M};\bm{V}^t]$ as input, $f_{\theta}$ predicts the features of all masked positions in a bi-directional manner, denoted as
$\widehat{\bm{V}}^t=p_{\theta}([\bm{M};\bm{V}^t])$, where $\widehat{\bm{V}}^{t}=[\hat{\bm{v}}_j^t] \in \mathbb{R}^{J \times D}$.

These predicted features are then compared with all movie shots to model the conditional probabilities of shots, \ie, for $i=1,...,I$ and $j=1,...,J$, 
\begin{eqnarray}\label{eq:sim}
\begin{aligned}
    &\widehat{\bm{S}}^t=[\hat{s}_{j,i}^t],~\text{with}~\hat{s}_{j,i}^t=\frac{\langle\hat{\bm{v}}_j^t, \bm{m}_i\rangle}{\|\hat{\bm{v}}_j^t\|_2\|\bm{m}_i\|_2},\\
    &\bm{P}^t = [p_{j,i}^t]=\text{Softmax}(\widehat{\bm{S}}_t)\in \mathbb{R}^{J\times I},
\end{aligned}
\end{eqnarray}
where $\widehat{\bm{S}}^t$ is the cosine similarity matrix. 
Passing it through a softmax operation leads to the conditional probability matrix $\bm{P}^t$, where $p_{j,i}^t$ indicates the probability of putting the $i$-th movie shot into the $j$-th position of the target trailer.


Given a set of movie-trailer pairs, denoted as $\mathcal{D}$, we train $f_{\theta}$ by maximizing the conditional likelihood of the predicted trailer shots, which is equivalent to minimizing a cross-entropy loss, \ie,
\begin{eqnarray}\label{eq:loss}
\begin{aligned}
    \min_{\theta}-\mathbb{E}_{\bm{M},\bm{V}\sim\mathcal{D}, t\sim\text{Unif}[0,1]}\Bigg[\frac{1}{t}\sum_{i=1}^{I}\sum_{j\in\mathcal{J'}}g_{j,i}\log p_{j,i}\Bigg].
\end{aligned}
\end{eqnarray}
Unlike the select-and-ranking methods~\cite{xu2015trailer,wang2024inverse,zhu2025weakly} and the autoregressive ones~\cite{argaw2024towards,jin2025tvmtrailer}, our model achieves bi-directional contextual modeling of movie trailer. 
The model is optimized using the AdamW optimizer~\cite{loshchilov2017decoupled}, performing iterative updates until convergence.

\subsection{A Self-Paced Mask Ratio Scheduler}
\label{sec:self-paced}

The mask ratio $t$ is critical for the model training, as it determines how much information is hidden from the model and thus reflects the reconstruction difficulty.
A low mask ratio results in a trivial task inadequate for model training, while a high mask ratio presents an overly difficult task that prevents the model from converging.
Inspired by the self-paced learning strategies~\cite{NIPS2010_e57c6b95,jiang2014self}, we design a self-paced masking strategy that adaptively schedules the mask ratio (\ie, task difficulty) to improve learning efficiency.

Specifically, we adjust the mask ratio in different optimization steps, leading to a mask ratio sequence, \ie, $\{t_1,t_2,...\}$. 
In the $n$-th optimization step, we compute the training loss in~\cref{eq:loss} for a batch of movie-trailer pairs (denoted as $\mathcal{B}$) and obtain the training accuracy as
\begin{eqnarray}\label{eq:acc}
\begin{aligned}
    a_n&=\frac{1}{|\mathcal{B}|}\sideset{}{_{\{\bm{M},\bm{V}\}\in\mathcal{B}}}\sum\left[\frac{1}{|\mathcal{J}'|}\sideset{}{_{j\in\mathcal{J}'}}\sum\mathbb{I}[g_{j,i'}=1]\right],\\
    i'&=\arg\sideset{}{_{i\in\{1,...,I\}}}\max p^{t_n}_{j,i},
\end{aligned}
\end{eqnarray}
where $\mathbb{I}[\cdot]$ is the indicator outputting $1$ if the input statement is true and $0$, otherwise.

The training accuracy indicates how well the model handles the current training task. 
At the early stage of training, we need to set a low mask ratio for a warm start.
When the accuracy increases rapidly in a few optimization steps, it indicates that the current task is too simple to train the model sufficiently, and we should increase the mask ratio.
On the contrary, when the accuracy is low given the current mask ratio, it means that the current task is difficult enough, and we should maintain the mask ratio until the model is well-trained.
Note that, to encourage our method to improve the model itself rather than retreating to easier tasks, we do not reduce the mask ratio during training.



Motivated by the above analysis, we adjust the mask ratio according to the current and historical training accuracy, proposing the following momentum-based scheduler:
\begin{eqnarray}\label{eq:self-paced}
\begin{aligned}
    b_{n+1} &= \mu_{a}a_{n} + (1 - \mu_{a})b_{n},\\
    \tilde{t}_{n+1} &= \mu_{t}t_{n} + (1 - \mu_{t})[t_{\min} + \Delta t\cdot\sigma_{\beta}(b_{n+1}-0.5)],\\
    t_{n+1} &=\max\{t_{n},~\tilde{t}_{n+1}\}.
\end{aligned}
\end{eqnarray}
In the first line of~\cref{eq:self-paced}, $b_{n+1}$ is the momentum term indicating the impact of historical training accuracy till the $n$-th step.
$\mu_a\in [0,1]$ works for balancing the current training accuracy $a_n$ and the previous momentum $b_n$.
In the second line of~\cref{eq:self-paced}, $t_{\min}$ denotes the predefined minimal mask ratio, and $\Delta t=t_{\max}-t_{\min}$ denotes the range of mask ratio.
In this study, we set $t_{\min}=0.1$ and $t_{\max}=1$, respectively.
$\sigma_{\beta}(b_{n+1}-0.5)=\frac{1}{1+\exp(-\beta(b_{n+1}-0.5))}$ achieves a smooth nonlinear map of the momentum term. 
Accordingly, $t_{\min} + \Delta t\cdot\sigma_{\beta}(b_{n+1}-0.5)$ works to adjust the mask ratio $t_n$, which approaches to $t_{\min}$ or $t_{\max}$, respectively, when $b_{n+1}$ is too low or too high. 
$\mu_t\in [0,1]$ determines the significance of the adjustment.
Finally, the third line of~\cref{eq:self-paced} ensures the mask ratio is monotonously increasing. 
This constraint is reasonable --- after the model has done well at the current difficulty level, we shall not reduce the task difficulty in the following optimization steps.


Leveraging the scheduler in~\cref{eq:self-paced}, the mask ratio increases in a self-paced way, enabling the model to learn from challenging tasks efficiently based on its current performance.
When accuracy fluctuates significantly, the momentum terms stabilize the update by preventing the mask ratio from changing excessively.


\subsection{Trailer Generation with Self-Correction}
\label{sec:3_3}

Given the trained model $f_{\theta}$, we design a new trailer generation process with a progressive self-correction mechanism.
Given a movie, we segment it into shots~\cite{soucek2024transnet} and extract its shot features $\bm{M}\in\mathbb{R}^{I\times D}$ by the ImageBind-based video encoder~\cite{girdhar2023imagebind}.
To generate a trailer with $J$ shots,\footnote{Similar to the work in~\cite{zhu2025weakly}, we determine the number of trailer shots based on the music track associated with the trailer.
We set $J$ to match the number of music segments detected by Ruptures~\cite{truong2020selective}.} we initialize a shot sequence $[\bm{M};\bm{V}_0]$, where $\bm{V}_0 = [\text{MP}] \in \mathbb{R}^{J \times D}$ is a fully masked sequence, and predict the masked features by $f_{\theta}$.
In addition, we introduce a vector $\bm{q}=[q_j]\in [0,1]^J$, in which $q_j$ indicates whether a masked position should be filled.
This vector is initialized as a zero vector.
In principle, our model achieves assigning $J$ distinct movie shots to the masked positions iteratively.


As illustrated in \cref{fig:inference}, the model predicts all masked features iteratively, \ie, $\widehat{\bm{V}}_{k}=[\hat{\bm{v}}_{j,k}]=f_{\theta}([\bm{M};\bm{V}_k])$, for $k=0,1,2....$. 
In the $k$-th iteration, the set of remaining unselected movie shots and that of the masked positions are denoted as $\mathcal{I}_k$ and $\mathcal{J}_k$, respectively.
Given $\widehat{\bm{V}}_{k}$ and $\bm{M}$, we compute the similar matrix and the conditional probability matrix by~\cref{eq:sim}.
Specifically, for $i\in\mathcal{I}_k$ and $j=1,...,J$, we have
$\hat{s}_{j,i}=\frac{\langle\hat{\bm{v}}_{j,k},\bm{m}_j\rangle}{\|\hat{\bm{v}}_{j,k}\|_2\|\bm{m}_j\|_2}$.
Given, $\widehat{\bm{S}}=[\hat{s}_{j,i}]$, we set $\bm{P}=[p_{j,i}]=\text{Softmax}(\widehat{\bm{S}})$.
According to $\bm{P}$, we assign a candidate movie shot to each masked position and update the mask-filling probability: for $j=1,...,J$, 
\begin{eqnarray}\label{eq:pos}
\begin{aligned}
    i_j^* = \arg\sideset{}{_{i
    \in\mathcal{I}_k}}\max p_{j,i},\quad
    q_j = \min\{q_j + p_{j,i_j^*},~1\}.
\end{aligned}
\end{eqnarray}
Here, we ensure $q_j$ is monotonously increasing.
For each candidate shot $\bm{m}_{i_j^*}$, the probability $q_j$ indicates how confidence our model assign the shot to the $j$-th masked position. 
Therefore, we only sample and maintain some high-confidence shots randomly and re-mask the remaining low-confidence ones.
In particular, each selected movie shot is inserted into $\bm{V}_k$, leading to the model input $\bm{V}_{k+1}=[\bm{v}_{j,k+1}]$ in the next iteration: for $j=1,...,J_k$, we sample $\tau\sim \text{Bernoulli}(q_j)$ and set
\begin{eqnarray}\label{eq:remask}
\begin{aligned}
    \bm{v}_{j,k+1}=
    \begin{cases}
    \bm{m}_{i_j^*}, & \text{if $\tau=1$ and $\bm{v}_{j,k}=\text{MP}$},\\
    \bm{v}_{j,k}, & \text{if $\tau=1$ and $\bm{v}_{j,k}\neq\text{MP}$},\\
    \text{MP}, & \text{otherwise.}
    \end{cases},
\end{aligned}
\end{eqnarray}
Based on the assignment results, we remove the selected movie shots from the candidate set $\mathcal{I}$ to prevent duplicate selection in subsequent steps.
Similarly, the filled position is removed from the position set $\mathcal{J}$.
Note that, our trailer generation process can converge in finite iterations --- with the increase of iterations, all $q_j$'s become $1$ gradually and thus all the masked positions are definitely filled.

\cref{alg:self-correction} shows the scheme of our method in details.
In our method, the shots selected in the current iteration can be re-masked in the next iteration. 
This process achieves a self-correction of the current prediction results and thus mitigate the error propagation issue, which makes our method imitate how human editors work (\ie, adjusting the shots at different positions iteratively).




\begin{algorithm}[t]
\caption{Self-Corrective Trailer Generation}\label{alg:self-correction}
\begin{algorithmic}[1]
    \STATE \textbf{Input:} Trained model $f_{\theta}$, movie shots $\bm{M}=[\bm{m}_i]\in\mathbb{R}^{I\times D}$, and the trailer length $J$. 
    \STATE \textbf{Output:} A trailer index vector $\bm{z}=[z_j]$.
    \STATE Initialize $k=0$, $\bm{z} =\bm{q}= \bm{0}_J$, $\mathcal{I}_k=\{1,...,I\}$, $\mathcal{J}_k=\{1,...,J\}$, and $\bm{V}_k=[\text{MP}]\in\mathbb{R}^{J\times D}$.
    \WHILE{$\mathcal{J}_k\neq\emptyset$}
        \STATE Predict $\widehat{\bm{V}}_{k}=[\hat{\bm{v}}_{j,k}]=f_{\theta}([\bm{M};\bm{V}_k])$.
        \STATE Set $\bm{V}_{k+1}=\bm{V}_{k}$, $\mathcal{I}_{k+1}=\mathcal{I}_{k}$, and $\mathcal{J}_{k+1}=\mathcal{J}_{k}$.
        \STATE Obtain $\bm{P}=[p_{j,i}]$ by \cref{eq:sim}, $i,j\in\mathcal{I}_k\times\{1,..,J\}$.
        \FOR{$j=1,...,J$}
            \STATE Obtain $i_j^*$ and update $q_j$ by~\cref{eq:pos}.
            \STATE $\tau\sim\text{Bernoulli}(q_j)$. Update $\bm{v}_{j,k+1}$ by~\cref{eq:remask}.
            \IF{$\tau=1$}
                \STATE $\mathcal{I}_{k+1}=\mathcal{I}_{k+1}\setminus\{i_j^*\}$, $\mathcal{J}_{k+1}=\mathcal{J}_{k+1}\setminus\{j\}$, and $z_j=i_j^*$ if $\bm{v}_{j,k}=\text{MP}$.
            \ELSE           
                \STATE $\mathcal{I}_{k+1}=\mathcal{I}_{k+1}\cup\{z_j\}$, $\mathcal{J}_{k+1}=\mathcal{J}_{k+1}\cup\{j\}$, and $z_j=0$ if $\bm{v}_{j,k}\neq\text{MP}$.
            \ENDIF
        \ENDFOR
        \STATE $k=k+1$.
    \ENDWHILE
    \RETURN $\bm{z}$.
\end{algorithmic}
\end{algorithm}

\subsection{Post-Processing for Trailer Generation}\label{ssec:post}
Based on the trailer index vector $\bm{z}$ obtained by~\cref{alg:self-correction}, we select the corresponding shots from the given movie and generate a trailer, denoted as $\mathcal{T}_{\bm{z}}$.
Following the work in~\cite{zhu2025weakly}, we adopt the music track from the official trailer of the movie for $\mathcal{T}_{\bm{z}}$ and perform the corresponding post-processing steps. 
We first segment the music track by Ruptures~\cite{truong2020selective} and align the duration of each trailer shot with its corresponding music shot.
We then employ DeepSeek-V3~\cite{deepseekai2024deepseekv3technicalreport} to analyze and select key subtitles as trailer narrations and apply MiniCPM-V2.6~\cite{yao2024minicpm} to generate one-sentence descriptions for each trailer shot.
To determine the shot positions of the narrations, we utilize CLIP~\cite{radford2021learning} to extract textual representations of narrations and shot descriptions and align them by maximizing their textual similarity using dynamic programming~\cite{bellman1966dynamic}.
Please refer to the work in~\cite{zhu2025weakly} or our supplementary file for more details.


\begin{table*}[t]
\caption{Numerical results compared with methods of different categories. 
The best results are bold and the second-best are underlined.}
\label{tab:main_table}
\setlength{\abovecaptionskip}{8pt}
\centering
\tabcolsep=3.8pt
\small{
\begin{tabular}{c|c|ccccc|ccccc}
\toprule
\multirow{2}{*}{Category} &
\multirow{2}{*}{Method} & 
\multicolumn{5}{c|}{Test-8} &
\multicolumn{5}{c}{Test-74} \\
\cline{3-12}
&
&Precision$\bm{\uparrow}$   &Recall$\bm{\uparrow}$  &F1$\bm{\uparrow}$ 
&LD$\bm{\downarrow}$   &AA$\bm{\uparrow}$   &Precision$\bm{\uparrow}$  &Recall$\bm{\uparrow}$    &F1$\bm{\uparrow}$  &LD$\bm{\downarrow}$   & AA$\bm{\uparrow}$\\ \hline
&VASNet~\cite{fajtl2018summarizing}                    
&0.0712     &0.0645     &0.0676     &100.62     &0.43
&0.0496     &0.0422     &0.0455     &84.79     &0.44  \\
Video&CLIP-It~\cite{narasimhan2021clip} 
&0.0711     &0.0629     &0.0667     &101.12     &0.45
&0.0832     &0.0710     &0.0764     &85.82     &0.38  \\
Summarization&OTVS~\cite{wang2023self}
&0.0688     &0.0613     &0.0648     &101.25     &0.46
&0.0834     &0.0711     &0.0766     &84.82     &0.39  \\
&Muvee~\cite{ganhor2014muvee}
&\textbf{0.2400}     &0.0461     &0.0714     &103.50     &0.36
&--     &--     &--     &--     &--  \\
\hline
&M2T~\cite{rehusevych2019movie2trailer}
&0.0611     &0.0503     &0.0515     &\textbf{95.67}     &0.42
&--     &--     &--     &--     &--  \\
&V2T~\cite{irie2010automatic}
&0.1121     &0.0603     &0.0945     &103.75     &0.52
&--     &--     &--     &--     &--  \\
Selection-&PPBVAM~\cite{xu2015trailer}
&0.0813     &0.1244     &0.0945     &101.50     &0.53
&--     &--     &--     &--     &--  \\
then-Ranking&IPOT~\cite{wang2024inverse}
&0.1218     &0.1425 &0.1311     &101.50     &0.42
&0.1258   &0.1483    &0.1357     &83.21     &0.46  \\
&MMSC~\cite{zhu2025weakly}
&0.1301     &\underline{0.1449}     &\underline{0.1391}     &\underline{99.25}     &\underline{0.58}
&\underline{0.1852}     &\underline{0.2164}     &\underline{0.1991}
&\underline{82.48}     &\underline{0.50}  \\
\hline
Autoregression&TGT~\cite{argaw2024towards}
&0.1185     &0.1142     &0.1153     &103.87     &0.48
&0.1243     &0.1433     &0.1326     &88.29     &0.43  \\
\hline
\rowcolor[HTML]{F2F2F2}
Masked Prediction& SSMP(Ours)
&\underline{0.1563}     &\textbf{0.1679}     &\textbf{0.1618}     &99.50   &\textbf{0.68}
&\textbf{0.2212}     &\textbf{0.2573}     &\textbf{0.2373}     &\textbf{81.87}     &\textbf{0.67}  \\
\bottomrule
\end{tabular}
}
\vspace{-8.0pt}
\end{table*}
\section{Experiment}
\label{sec:experiment}

To demonstrate the effectiveness of our SSMP method, we conduct both objective and subjective evaluations compared with state-of-the-art methods.

\subsection{Implementation Details}
\textbf{Datasets.}
We extend the CMTD dataset~\cite{wang2024inverse} to train our model and evaluate the model on the two testing sets of the dataset, called Test-8 and Test-74, respectively.
This dataset contains 500 movies and 922 corresponding trailers.
We further enrich 30 movie-trailer pairs newly released in 2024 to construct a new test set, called Test-2024, which is used to further evaluate the model’s generalization ability.
The collection of all three testing sets is called Test-ALL, which is applied in ablation studies.
For each video (a movie or a trailer), we segment shots and extract their features following the pre-processing pipeline in~\cite{wang2024inverse,sun2025multi}.

\textbf{Baselines.}
We compare our SSMP method with state-of-the-art trailer generation methods, including selection-then-ranking methods (\eg, V2T~\cite{irie2010automatic}, M2T~\cite{rehusevych2019movie2trailer}, PPBVAM~\cite{xu2015trailer}, IPOT~\cite{wang2024inverse}, and MMSC~\cite{zhu2025weakly}) and the autoregressive method TGT\footnote{Since the data and code are not publicly available, we reproduce TGT~\cite{argaw2024towards} by ourselves.
When generating a trailer, we set the same length for TGT and our method for a fair comparison.}~\cite{argaw2024towards,jin2025tvmtrailer}.
Here, IPOT~\cite{wang2024inverse} and MMSC~\cite{zhu2025weakly} require background music as input, while the others only leverage visual information.
Several video summarization methods are considered when evaluating the shot selection performance, including Muvee~\cite{ganhor2014muvee}, VASNet~\cite{fajtl2018summarizing}, CLIP-It~\cite{narasimhan2021clip}, and OTVS~\cite{wang2023self}. 
For a fair comparison, all the methods are tested on the same dataset, and all the trailer generation methods apply the post-processing shown in \cref{ssec:post}.

\textbf{Evaluation metrics.}
To evaluate the shot selection performance, we employ Precision, Recall, and F1-score (F1) by comparing each $\bm{z}$ with the ground-truth movie–trailer alignment.
Following the work in~\cite{liu2023video, argaw2024towards,jin2025tvmtrailer}, we further use Levenshtein Distance (LD) and Pairwise Agreement Accuracy (AA) to measure the shot ranking performance.
For each generated trailer and the official one, LD quantifies their dissimilarity by computing the minimum number of edits required to convert one shot sequence into another, and AA measures the proportion of well-aligned shot pairs.

\textbf{Hyperparameter settings.}
Our model has four Transformer layers, and each layer has four attention heads with a hidden size of 1,024 and a feed-forward size of 2,048.
The feature dimension $D$ is 1,024.
The hyperparameters of the self-paced mask ratio scheduler are empirically set to $\beta=10$, $\mu_a = 0.98$, and $\mu_t = 0.1$.
For the AdamW optimizer, we set a learning rate of $10^{-4}$ under a cosine scheduler and a warmup ratio of $0.1$.
We set the weight decay to $0.1$, $\beta_2$ to $0.95$.
Our model is trained for 500 epochs with a batch size of 5, on a single NVIDIA H100.

\begin{table}[t]
\caption{Numerical results on newly released movies.}
\label{tab:test-2024}
\setlength{\abovecaptionskip}{8pt}
\centering
\tabcolsep=4pt
\small{
\begin{tabular}{c|ccccc}
\toprule
\multirow{2}{*}{Method} & 
\multicolumn{5}{c}{Test-2024}\\
\cline{2-6}
&Precision$\bm{\uparrow}$  &Recall$\bm{\uparrow}$   &F1$\bm{\uparrow}$ &LD$\bm{\downarrow}$  &AA$\bm{\uparrow}$\\
\hline
VASNet~\cite{fajtl2018summarizing}         
&0.0990     &0.0806    &0.0886
&93.84      &0.44\\
CLIP-It~\cite{narasimhan2021clip} 
&0.0443    &0.0370     &0.0403
&95.51     &0.31\\
OTVS~\cite{wang2023self}
&0.1021   &0.0831     &0.0914
&\underline{93.56}    &0.44\\
\hline
IPOT~\cite{wang2024inverse}
&0.1236    &0.1460     &0.1336
&93.82     &0.42\\
MMSC~\cite{zhu2025weakly}
&\underline{0.1567}     &0.1485     &0.1506
&93.91      &\underline{0.52}\\
\hline
TGT~\cite{argaw2024towards}
&0.1521     &\underline{0.1601}     &\underline{0.1601}
&103.25      &0.46\\
\hline
\rowcolor[HTML]{F2F2F2}
SSMP(Ours)
&\textbf{0.1587}     &\textbf{0.1999}     &\textbf{0.1759}
&\textbf{93.29}     &\textbf{0.60}\\ 
\bottomrule
\end{tabular}
}
\vspace{-10.0pt}
\end{table}

\begin{figure*}[t]
    \centering
    \begin{subfigure}[t]{0.725\textwidth}
        \centering
        \includegraphics[width=\textwidth]{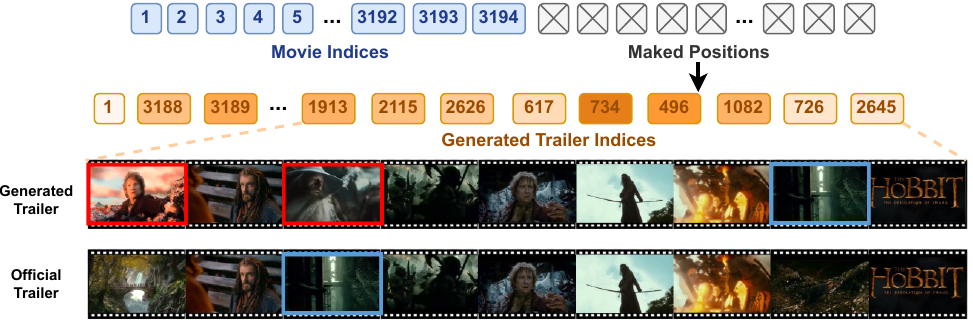}
        \caption{Comparison between generated and official trailer shots of \textit{The Hobbit}}
        \label{fig:visual}
    \end{subfigure}
    \hfill
    \begin{subfigure}[t]{0.265\textwidth}
        \centering
        \includegraphics[width=\textwidth]{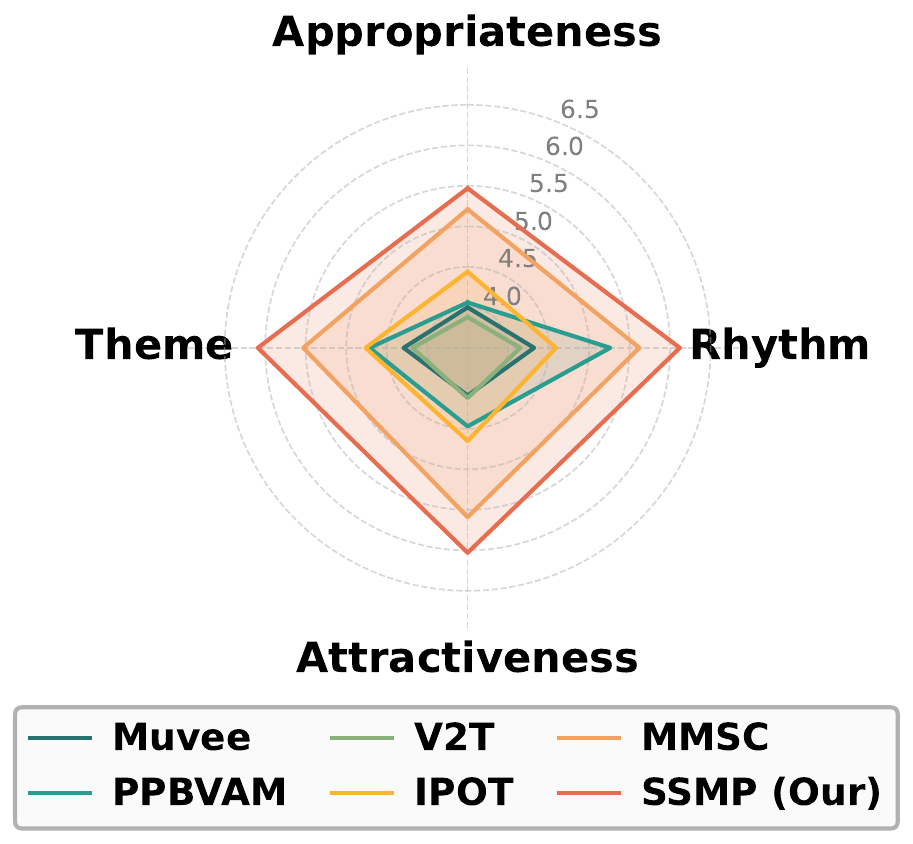}
        \caption{Subjective scores}
        \label{fig:radar}
    \end{subfigure}
    \caption{(a) Red boxes mark incorrect predictions, and blue boxes highlight the predicted and real positions of a correctly selected shot. (b) Radar chart illustrating the average scores of different methods across four aspects.}
    \label{fig:scheme}
    \vspace{-10.0pt}
\end{figure*}

\subsection{Comparisons}
\textbf{Quantitative comparisons.} \cref{tab:main_table} shows the experimental comparison between \textbf{SSMP} and baselines.
SSMP achieves the best performance on most metrics by a substantial margin.
Specifically, the F1 scores surpass the state-of-the-art selection-then-ranking method (\ie, MMSC) by 2.27\% and 3.82\% on the two test sets, respectively. 
The AA metric exhibits significant improvements of 10\% and 17\% on the two test sets, demonstrating the superior capability of our method in modeling temporal dependencies compared with the methods that decouple shot selection and ranking.
Besides, unlike the autoregressive method (\ie, TGT) that generates shots strictly in a sequential manner, our method flexibly fills the most confident position and re-masks the others at each step, which enables the model to reconsider and correct uncertain predictions based on global context, thereby reducing the impact of uncertain predictions.
Furthermore, the results in \cref{tab:test-2024} demonstrate that our method generalizes well to newly released movies.


\textbf{Qualitative comparisons.}
\cref{fig:visual} visualizes a generated trailer and the official trailer.
Each shot of the generated trailer is labeled by the shot index in the original movie. 
The colors of the indices reflect the number of iterations required to generate the shots.
The darker color a shot has, the more iterations are used to generate it.
As aforementioned, this iterative generation process allows self-correction --- the subsequent predictions are conditioned on increasingly reliable contextual representations, and accordingly, error propagation is suppressed.
As a result, the generated trailers exhibit higher ordering accuracy (\ie, AA), which aligns with the quantitative results reported in \cref{tab:main_table}.

To evaluate the perceptual quality of the generated trailers, we conduct a user study involving 25 participants.
Each participant compares the trailers generated by our method and several baselines with the corresponding official trailers, where all trailers are anonymous to ensure unbiased evaluation.
Following in the work in~\cite{irie2010automatic,wang2024inverse,zhu2025weakly}, the evaluation is performed from four aspects: \textbf{theme} (\ie, how well the trailer conveys the theme of the movie), \textbf{rhythm} (\ie, how well the visuals align with the rhythm of the music), \textbf{attractiveness} (\ie, how attractive the trailer appears to the viewer), and \textbf{appropriateness} (\ie, how close the trailer is to the official trailer), which together provide a comprehensive assessment of the perceived gap between generated and official trailers.
The score of each aspect ranges from 1 to 7, with higher scores corresponding to better performance.
The radar chart in \cref{fig:visual} illustrates the average scores of all participants and demonstrates that our method consistently outperforms previous methods.


\subsection{Ablation Studies}
\begin{figure}[t]
    \centering
    \begin{subfigure}[t]{0.49\linewidth}
        \centering
        \includegraphics[width=\linewidth]{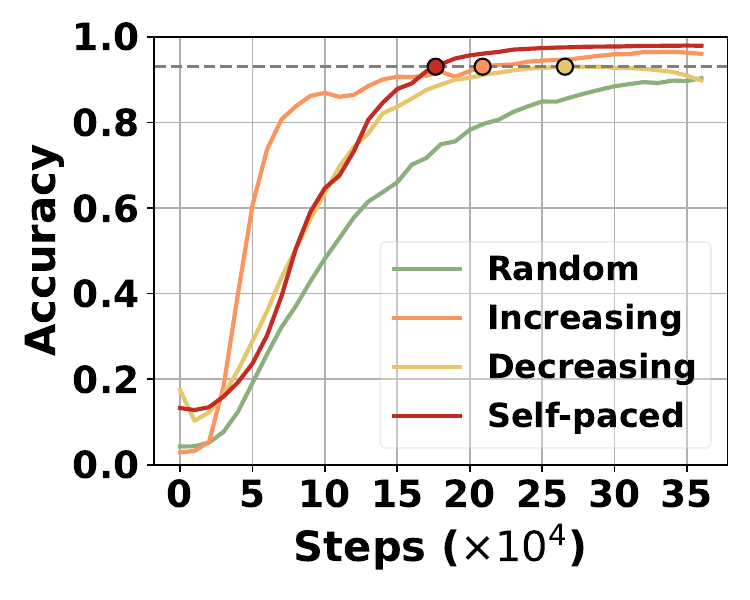}
        \caption{accuracy-step}
        \label{fig:acc_steps}
    \end{subfigure}
    \hfill
    \begin{subfigure}[t]{0.49\linewidth}
        \centering
        \includegraphics[width=\linewidth]{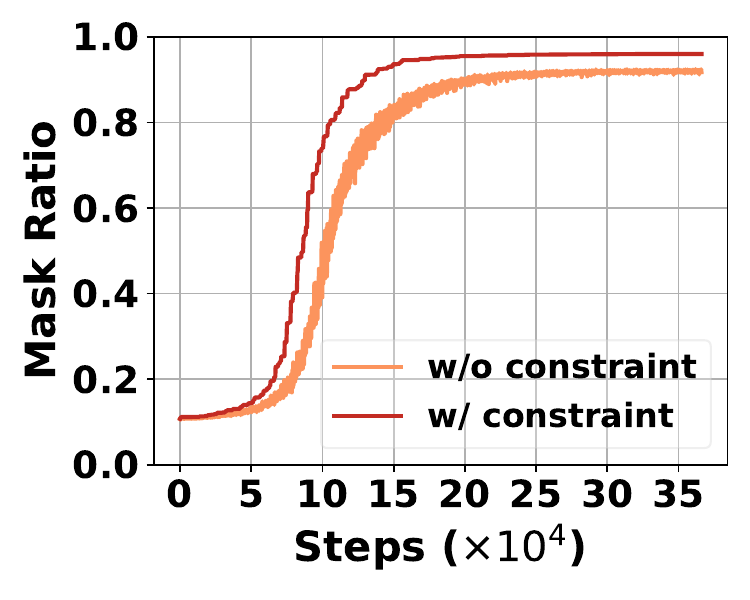}
        \caption{ratio-step}
        \label{fig:ratio_steps}
    \end{subfigure}
    \caption{Mask ratio and prediction accuracy over optimization steps. Ratios in (b) are averaged over every 100 steps.}
    \label{fig:merged_line}
\end{figure}

\textbf{Mask Ratio Setting.}
When training our model, we explore four mask ratio schedulers: a random mask ratio, a linearly increased mask ratio, a linearly decreased mask ratio, and the proposed self-paced strategy.
As shown in \cref{tab:self-paced}, gradually increasing the mask ratio allows the model to start with easy tasks and progressively adapt to difficult ones, resulting in more stable optimization and better performance.
Moreover, our self-paced strategy achieves the best performance by adjusting the mask ratio based on the model performance during training, which verifies the rationality of our method.
We further visualize the training accuracy achieved by different mask ratio schedulers in \cref{fig:acc_steps}.
The points are the intersections between the gray dashed line and the curves of different strategies, representing the optimization steps required to achieve an accuracy of $0.95$.
It can be observed that our self-paced learning converges faster and achieves a higher upper bound than the other strategies.

\cref{tab:momentum} shows the results with different values of the hyperparameter $\mu_t$, which controls the update momentum of the mask ratio during training.
The results demonstrate that $\mu_t = 0.1$ provides a trade-off between stability and adaptability in the self-paced masking.
Note that, our self-paced scheduler is robust to the setting of the hyperparameter, with performance remaining stable for $\mu_t \in [0.1,0.9]$.
\begin{table}[t]
\caption{The impacts of different mask ratio schedulers.}
\label{tab:self-paced}
\setlength{\abovecaptionskip}{8pt}
\centering
\tabcolsep=3pt
\small{
\begin{tabular}{c|ccccc}
\toprule
\multirow{2}{*}{Ratio Setting} & 
\multicolumn{5}{c}{Test-ALL}\\
\cline{2-6}
&Precision$\bm{\uparrow}$  &Recall$\bm{\uparrow}$   &F1$\bm{\uparrow}$ &LD$\bm{\downarrow}$  &AA$\bm{\uparrow}$\\
\hline
Random                 
&0.1284     &0.1489    &0.1377
&\underline{95.25}      &0.64\\
Linearly decreased
&0.1643    &0.1901     &0.1760
&92.57     &\underline{0.66}\\
Linearly increased
&\underline{0.1791}    &\underline{0.2064}     &\underline{0.1915}
&92.55    &\underline{0.66}\\
\rowcolor[HTML]{F2F2F2}
Self-paced
&\textbf{0.1888}     &\textbf{0.2126}     &\textbf{0.1996}
&\textbf{91.19}     &\textbf{0.68}\\ 
\bottomrule
\end{tabular}
}
\end{table}

\begin{table}[t]
\caption{The robustness test of mask ratio momentum.}
\label{tab:momentum}
\setlength{\abovecaptionskip}{8pt}
\centering
\tabcolsep=3.5pt
\small{
\begin{tabular}{c|ccccc}
\toprule
\multirow{2}{*}{Ratio Momentum} & 
\multicolumn{5}{c}{Test-ALL}\\
\cline{2-6}
&Precision$\bm{\uparrow}$  &Recall$\bm{\uparrow}$   &F1$\bm{\uparrow}$ &LD$\bm{\downarrow}$  &AA$\bm{\uparrow}$\\
\hline
$\mu_t=0.9$                 
&0.1626     &0.1888    &0.1744
&93.00      &0.64\\
$\mu_t=0.5$ 
&\underline{0.1672}    &\underline{0.1933}     &\underline{0.1790}
&\underline{92.94}     &\underline{0.66}\\
\rowcolor[HTML]{F2F2F2}
$\mu_t=0.1$ 
&\textbf{0.1888}     &\textbf{0.2126}     &\textbf{0.1996}
&\textbf{91.19}     &\textbf{0.68}\\ 
$\mu_t=0.0$ 
&0.1632    &0.1898    &0.1753
&93.19     &0.67\\ 
\bottomrule
\end{tabular}
}
\vspace{-8.0pt}
\end{table}


\textbf{Monotonicity Constraint.}
As mentioned in~\cref{sec:self-paced}, we set a monotonicity constraint in the third line of ~\cref{eq:self-paced} to prevent the task difficulty from decreasing during training.
In~\cref{fig:ratio_steps}, without the monotonicity constraint, the mask ratio keeps fluctuating throughout the training process and causes slower convergence.
The final ratio reaches around $0.9$, indicating that the training task difficulty is limited and lower than in the monotonicity constraint setting.


\textbf{Self-Correction Mechanism.}
When generating a trailer, we explore the effectiveness of the proposed self-correction mechanism by comparing it with a greedy strategy that fills only the highest-confidence position at each iteration.
As shown in~\cref{tab:self-correction_mechanism}, applying self-correction leads to a better performance.
The confidence vector $\bm{q}$ in~\cref{eq:pos} represents accumulated confidence for each position, which smooths noise and identifies positions that consistently achieve high confidence.
Unlike the greedy strategy, this mechanism allows the low-confidence positions to be reconsidered and produce more reliable predictions across iterations.

\begin{table}[t]
\caption{The effectiveness of the self-correction mechanism.}
\label{tab:self-correction_mechanism}
\setlength{\abovecaptionskip}{8pt}
\centering
\tabcolsep=4.2pt
\small{
\begin{tabular}{c|ccccc}
\toprule
\multirow{2}{*}{Strategy} & 
\multicolumn{5}{c}{Test-ALL}\\
\cline{2-6}
&Precision$\bm{\uparrow}$  &Recall$\bm{\uparrow}$   &F1$\bm{\uparrow}$ &LD$\bm{\downarrow}$  &AA$\bm{\uparrow}$\\
\hline
Greedy       
&0.1879     &0.2055    &0.1958
&92.60      &0.67\\
\rowcolor[HTML]{F2F2F2}
Self-corrective
&\textbf{0.1888}     &\textbf{0.2126}     &\textbf{0.1996}
&\textbf{91.19}     &\textbf{0.68}\\ 
\bottomrule
\end{tabular}
}
\vspace{-10.0pt}
\end{table}

\textbf{Loss Functions.}
Besides the cross-entropy (CE) loss in \cref{eq:loss}, we consider training our model by minimizing the mean squared error (MSE) between predicted and ground-truth shot embeddings (\ie, $\min_{\theta}\mathbb{E}[\frac{1}{t}\sum_{j\in\mathcal{J}'}\|\hat{\bm{v}}_j^t-\bm{m}_{\ell_j}\|]$, where $g_{j,\ell_j}=1$ for $j=1,...,J$). 
\cref{tab:losses} shows that the CE loss leads to a better model.
This improvement can be attributed to the intrinsic difference between the two losses. 
Specifically, CE aims to push up the score of the correct shot while suppressing the others, which enhances the model’s ability to capture inter-shot distinctions.
MSE relies on the distance between shots in a high-dimensional space, which suffers the curse-of-dimensionality severely and thus lacks a discriminative margin.

\begin{table}[t]
\caption{The impacts of different training losses.}
\label{tab:losses}
\setlength{\abovecaptionskip}{8pt}
\centering
\tabcolsep=5.5pt
\small{
\begin{tabular}{c|ccccc}
\toprule
\multirow{2}{*}{Loss} & 
\multicolumn{5}{c}{Test-ALL}\\
\cline{2-6}
&Precision$\bm{\uparrow}$  &Recall$\bm{\uparrow}$   &F1$\bm{\uparrow}$ &LD$\bm{\downarrow}$  &AA$\bm{\uparrow}$\\
\hline
MSE Loss               
&0.1141     &0.1297    &0.1211
&96.92      &0.56\\
\rowcolor[HTML]{F2F2F2}
CE Loss
&\textbf{0.1888}     &\textbf{0.2126}     &\textbf{0.1996}
&\textbf{91.19}     &\textbf{0.68}\\ 
\bottomrule
\end{tabular}
}
\end{table}



\textbf{Shot Deviation.}
Referring to the work in~\cite{argaw2024towards, wang2024inverse, sun2025multi}, we also conduct experiments on allowable positional deviation during trailer generation.
This design is motivated by the observation that adjacent shots in movies often belong to the same event and exhibit highly similar visual content, making them difficult to distinguish precisely.
To mitigate the potential bias caused by such local ambiguity and ensure a more reasonable evaluation of model performance, we relax the matching criterion by including the $R$ neighboring shots around each official trailer shot within the ground truth set.
The results in \cref{tab:deviation} show that allowing small positional deviations consistently improves performance, confirming that adjacent shots typically share overlapping semantics and that strict positional alignment is therefore not essential for assessing model performance.

\begin{table}[t]
\caption{The impact of relaxing the shot selection metric. $R$ denotes the radius of the allowable positional deviation.}
\label{tab:deviation}
\centering
\tabcolsep=2pt
\small{
\begin{tabular}{c|ccc|ccc}
\toprule
\multirow{2}{*}{} & 
\multicolumn{3}{c|}{Test-8} &
\multicolumn{3}{c}{Test-74} \\
\cline{2-7}
&Precision$\bm{\uparrow}$  &Recall$\bm{\uparrow}$   &F1$\bm{\uparrow}$ &Precision$\bm{\uparrow}$  &Recall$\bm{\uparrow}$   &F1$\bm{\uparrow}$ \\
\hline
$R=0$                 
&0.1563     &0.1679     &0.1618
&0.2212     &0.2573     &0.2373     \\
$R=1$
&0.3795     &0.3909     &0.3894 
&0.4139     &0.4761     &0.4421     \\
$R=2$ 
&0.6760     &0.6194     &0.6306
&0.5815     &0.6700     &0.6216     \\ 
\bottomrule
\end{tabular}
}
\vspace{-10pt}
\end{table}


\section{Conclusion and Future Work}
\label{sec:conclusion}
In this paper, we develop SSMP, a self-paced masked generation method, which achieves bi-directional contextual modeling for automatic movie trailer generation.
We apply a self-paced learning strategy to train the model and design a self-corrective generation process, which imitates how human editors work.
This method achieves encouraging performance in automatic movie trailer generation, outperforming existing methods that apply selection-then-ranking and auto-regressive paradigms.

\textbf{Limitations and Future Work.} Currently, our method merely relies on the visual information of movie and trailer shots, without incorporating acoustic and textual information (\ie, trailer audio or textual metadata).
Besides, the number of movie–trailer pairs used for training remains limited, which may restrict the model generalizability.
In future work, we plan to extend the SSMP method by integrating multi-modal information and expanding the dataset to further boost model performance.
In addition, we will explore the applications of our self-paced and self-corrective masked prediction method in other fields.

\section*{Acknowledgement}
Supported in part by the Natural Science Foundation of Beijing Municipality (Grant No. 4262030) and the Beijing Major Science and Technology Project under Contract no. Z251100008425002.

{
    \small
    \bibliographystyle{ieeenat_fullname}
    \bibliography{main}
}

\clearpage
\setcounter{page}{1}
\maketitlesupplementary

\section{Model Architecture}
Our SSMP model adopts a Transformer-based encoder architecture that leverages full self-attention across all shots.
Given an input sequence of shot features $\bm{X}=[\bm{x}_1,\bm{x}_2,...\bm{x}_T]$ as the initial hidden state $\mathbf{h}^{(0)}$, the model processes it through a stack of $L$ Transformer encoder blocks.
Each block consists of a full self-attention layer and a feed-forward layer, both surrounded by residual connections and normalized using RMSNorm.

At the $l$-th layer, the hidden representation $\mathbf{h}^{(l)}\in\mathcal{R}^{T\times D}$ is updated as:
\begin{equation}
\begin{aligned}
\mathbf{h}'^{(l)} &= \mathbf{h}^{(l)} + \text{MSA}\big(\text{RMSNorm}(\mathbf{h}^{(l)})\big), \\
\mathbf{h}^{(l+1)} &= \mathbf{h}'^{(l)} + \text{FFN}\big(\text{RMSNorm}(\mathbf{h}'^{(l)})\big),
\end{aligned}
\end{equation}
where $\text{MSA}(\cdot)$ denotes a multi-head self-attention mechanism formulated as:
\begin{equation}
\begin{aligned}
\mathbf{Q} = \mathbf{h}^{(l)} W_Q,\quad
\mathbf{K} = \mathbf{h}^{(l)} W_K,\quad
\mathbf{V} = \mathbf{h}^{(l)} W_V, \\
\text{MSA}(\mathbf{Q}, \mathbf{K}, \mathbf{V}) 
= \text{softmax}\!\left(
\frac{(\mathbf{Q}R_\theta)(\mathbf{K}R_\theta)^{\top}}{\sqrt{D_k}}
\right)\mathbf{V},
\end{aligned}
\end{equation}
where $R_\theta$ represents the rotary positional embedding matrix~\cite{su2024roformer} to encode relative positions, and $W_Q$, $W_K$ and $W_V$ are learnable matrices.
The feed-forward network (FFN) takes $\mathbf{z}=\text{RMSNorm}(\mathbf{h}'^{(l)})$ as input and applies a gated SiLU activation with a linear expansion and projection:
\begin{equation}
\text{FFN}(\mathbf{z}) 
= \text{Linear}_{\text{out}}\!\left(
\text{SiLU}\!\left(\text{Linear}_{\text{up}}(\mathbf{z})\right)
\right),
\end{equation}
where $\text{Linear}_{\text{up}}$: $\mathbb{R}^{D}\rightarrow\mathbb{R}^{2D}$, and $\text{Linear}_{\text{out}}$: $\mathbb{R}^{2D}\rightarrow\mathbb{R}^{D}$.
Referring to~\cite{nie2025large}, our model replaces standard layer normalization with RMS normalization:
\begin{equation}
\text{RMSNorm}(\mathbf{h}) 
= \frac{\mathbf{h}}{\sqrt{\tfrac{1}{D}\|\mathbf{h}\|_2^2 + \epsilon}} 
\odot \mathbf{g},
\end{equation}
where $\mathbf{g}$ is a learnable scaling vector and 
$\epsilon$ is a small constant ensuring numerical stability.
This encoder architecture allows each shot to attend to all others, enabling bidirectional context modeling.

\section{Post-Processing Details}
As mentioned in~\cref{ssec:post}, we insert the selected narrations into the generated trailer through solving a dynamic programming problem.
Formally, given the similarity matrix $\bm{C}=[c_{n,j}] \in \mathbb{R}^{N \times J}$ between $N$ narration features and $J$ trailer shot description features, the optimal alignment is obtained by maximizing the cumulative similarity under the constraint that each narration audio duration $L_j^{\text{shot}}$ must not exceed the duration of its corresponding trailer shot $L_n^{\text{nar}}$, \ie, if $L_j^{\text{shot}} \ge L_n^{\text{nar}}$, 
\begin{eqnarray}
\label{eq:dp}
\begin{aligned}
D_{n,j} =\max\{D_{n-1,\,j-1} + c_{n,j}, \;D_{n,\,j-1}\},
\end{aligned}
\end{eqnarray}
where $D_{n,j}$ denotes the maximum accumulated similarity between the first $n$ narrations and the first $j$ trailer shots.
We initialize $D_{0,*}=0$ and $D_{*,0}=-\infty$. 
The optimal alignment path is then obtained by tracing back from $\arg\max_j D_{N,j}$.
After inserting the background music and the narrations into the trailer $\mathcal{T}_{\bm{z}}$, we obtain the final generated trailer with coherent audiovisual composition.

\section{Detailed settings of the user study.}

We conduct the user study with 25 participants aged between 10 and 52. 
Each participant evaluate trailers for three movies used in [19, 49, 50] (\textit{The Hobbit 2}, \textit{The Wolverine}, and \textit{300: Rise of an Empire}). 
For each movie, participants watch and rate official trailer and six anonymized trailers generated by our method and five baselines.

\section{More Experimental Results}

\begin{figure}[htbp]
    \centering
    \vspace{-4.0pt}
    \includegraphics[width=\linewidth]{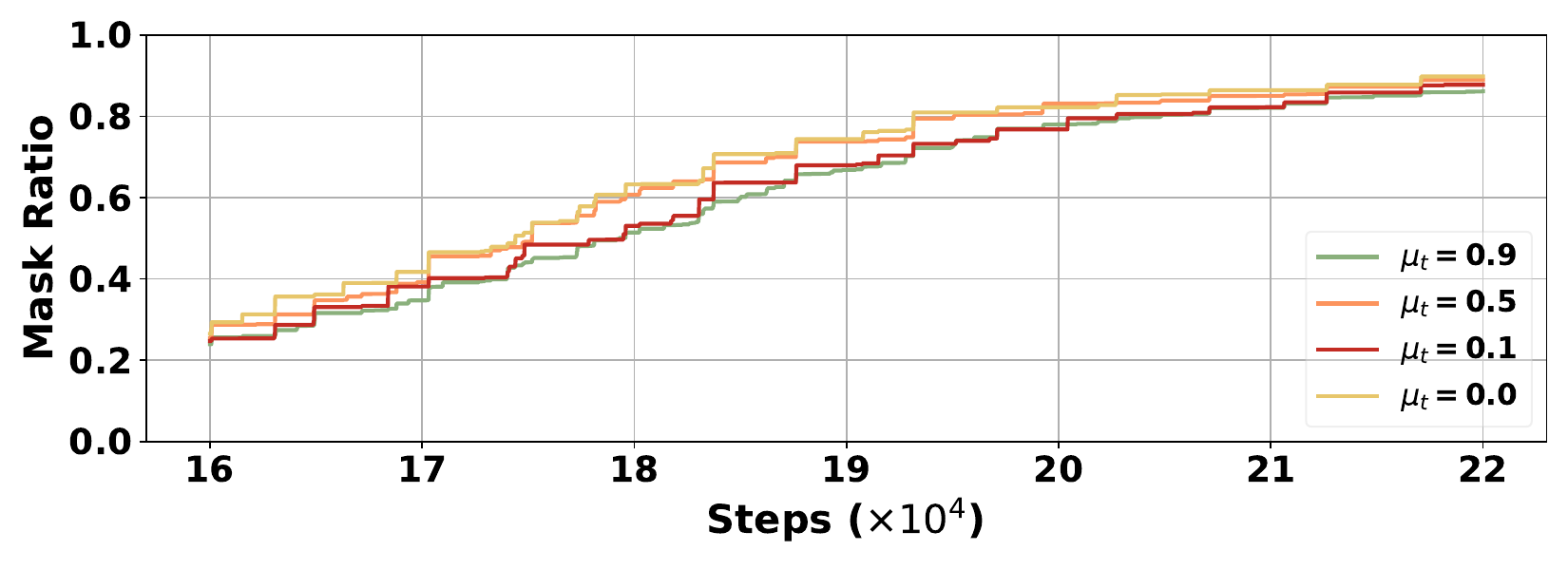}
    \vspace{-20.0pt}
    \caption{Mask ratio over training steps for different settings.}
    \label{fig:mask_ratio_momentum}
    \vspace{-4.0pt}
\end{figure}

\textbf{Mask Ratio Momentum.}
To further illustrate the results in~\cref{tab:momentum}, \cref{fig:mask_ratio_momentum} presents how the masking ratio changes during the middle of the training.
The curve of $\mu_t=0.1$ lies between those of $0.0$, $0.5$, and $0.9$, suggesting a balanced update of the mask ratio, achieving a trade-off between training efficiency and stability to accuracy changes.

\noindent\textbf{Training Accuracy Momentum.}
\cref{tab:acc_momentum} shows the results with different values of the hyperparameter $\mu_a$, which controls the balance between the current accuracy and its accumulated momentum during training.
The best results occur at $\mu_a = 0.98$, and the overall stability across different values further demonstrates its robustness.

\begin{table}[t]
\vspace{-4.0pt}
\caption{The robustness test of accuracy momentum.}
\label{tab:acc_momentum}
\vspace{-4.0pt}
\centering
\tabcolsep=2.5pt
\small{
\begin{tabular}{c|ccccc}
\toprule
\multirow{2}{*}{Accuracy Momentum} & 
\multicolumn{5}{c}{Test-ALL}\\
\cline{2-6}
&Precision$\bm{\uparrow}$  &Recall$\bm{\uparrow}$   &F1$\bm{\uparrow}$ &LD$\bm{\downarrow}$  &AA$\bm{\uparrow}$\\
\hline
$\mu_a=1.0$
&0.1661   &\underline{0.1925}    &0.1780
&93.06     &\underline{0.67}\\ 
\rowcolor[HTML]{F2F2F2}
$\mu_a=0.98 \text{(our)}$
&\textbf{0.1888}     &\textbf{0.2126}     &\textbf{0.1996}
&\textbf{91.19}     &\textbf{0.68}\\ 
$\mu_a=0.95$
&\underline{0.1743}   &0.1888    &\underline{0.1854}
&\underline{91.98}     &\textbf{0.68}\\ 
$\mu_a=0.9$
&0.1614   &0.1872    &0.1731
&92.90     &0.64\\ 
$\mu_a=0.5$
&0.1636   &0.1894    &0.1753
&93.34     &0.66\\ 
$\mu_a=0.0$
&0.1585   &0.1836    &0.1698
&93.34     &0.65\\ 
\bottomrule
\end{tabular}
}
\vspace{-10.0pt}
\end{table}


\noindent\textbf{Visual Encoder.}
Since previous trailer generation methods~\cite{wang2023self,zhu2025weakly} adopt ImageBind, we follow the same setting for a fair comparison and demonstrate that the performance gain is attributed by our method rather than by the encoder. 
Secondly, unlike image-based models (DINOv3 and SigLIP2), ImageBind provides a native visual encoder capturing spatio-temporal dynamics of shot, not requiring additional pooling of image features.
We replace ImageBind with SigLIP2 and DINOv3, learning shot-level representations under the same experimental setup.
\cref{tab:siglip2_dino3} shows that the three encoders are comparable in shot selection, verifying the rationality of using ImageBind. 

\begin{table}[ht]
\vspace{-4.0pt}
\caption{Testing different encoders on Test-ALL for shot selection.}
\label{tab:siglip2_dino3}
\vspace{-4.0pt}
\centering
\tabcolsep=3pt
\small{
\begin{tabular}{c|ccccc}
\toprule
Visual Encoder 
&Precision$\bm{\uparrow}$  &Recall$\bm{\uparrow}$   &F1$\bm{\uparrow}$ &LD$\bm{\downarrow}$  &AA$\bm{\uparrow}$\\
\hline
SigLIP2
&0.1873     &0.2018     &0.1936
&93.38      &0.67\\
DINOv3
&0.1894    &0.2106     &0.1990
&93.33     &0.66\\
\rowcolor[HTML]{F2F2F2}
ImageBind (Ours)
&0.1888    &0.2126    &0.1996
&91.19     &0.68\\ 
\bottomrule
\end{tabular}
}
\end{table}
\footnotetext[1]{SigLIP 2 adopts \texttt{siglip2-large-patch16-256}.}
\footnotetext[2]{DINOv3 adopts \texttt{dinov3-vitl16-pretrain-lvd1689m}.}
\vspace{-4.0pt}

\begin{figure}[h]
    \centering
    \begin{subfigure}[t]{0.49\linewidth}
        \centering
        \includegraphics[width=\linewidth]{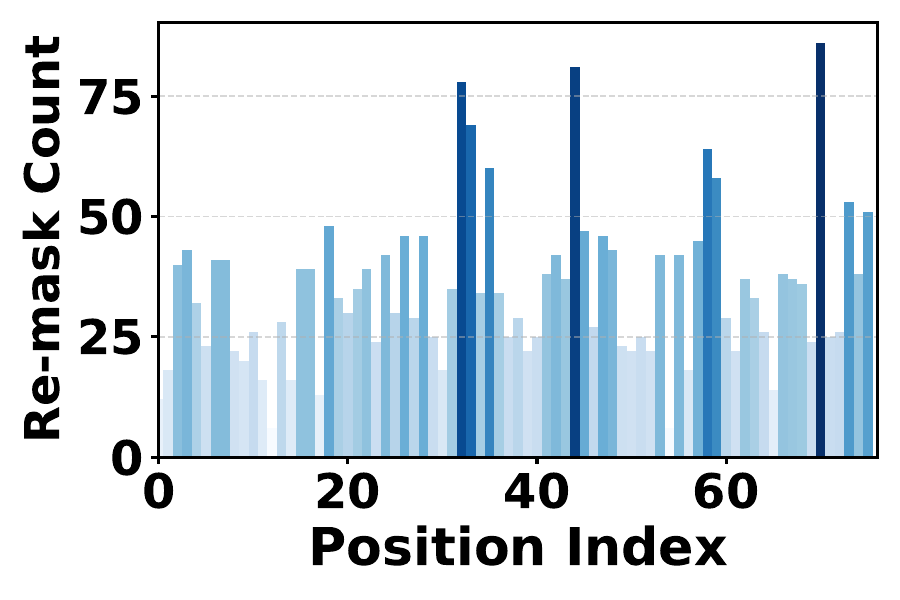}
        \label{fig:remask_count}
    \end{subfigure}
    \hfill
    \begin{subfigure}[t]{0.49\linewidth}
        \centering
        \includegraphics[width=\linewidth]{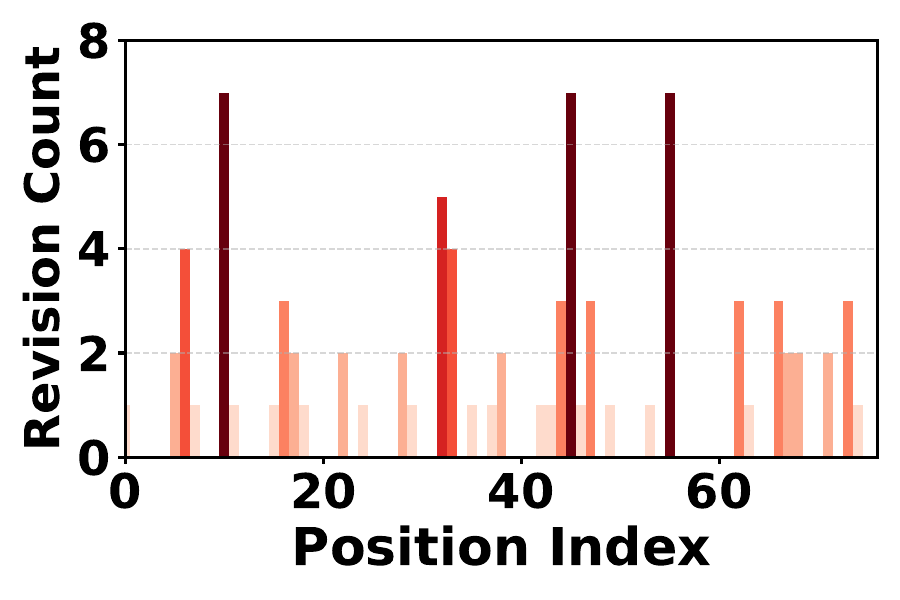}
        \label{fig:revision_count}
    \end{subfigure}
    \vspace{-15pt}
    \caption{Remasking and revision counts at each position index in a single trailer of 76 shots during the generation process.}
    \vspace{-4.0pt}
    \label{fig:remask_revision}
    
\end{figure}

\noindent\textbf{Self-correction Mechanism.}
\cref{fig:remask_revision} offers some insights into the self-correction across positions.
$i)$ The remasking and revision counts are unevenly distributed across positions, indicating that self-correction is selectively applied.
$ii)$ Most positions are re-masked at early stages.
As some positions become fixed, re-masking gradually concentrates on some key positions. 
$iii)$ The revision count is much lower than the remasking count --- remasking leads to a revision only when sufficient confidence is reached.
\textit{$iv)$ Notably, when applying a relaxed shot selection metric (e.g., allowable positional deviation=3), the superiority of self-correction becomes significant, as shown in~\cref{tab:R=3}.}
\vspace{-4.0pt}
\begin{table}[h]
\vspace{-4.0pt}
\caption{Numerical results under a relaxed shot selection metric.}
\label{tab:R=3}
\centering
\tabcolsep=0.5pt
\small{
\begin{tabular}{c|ccc|ccc}
\toprule
\multirow{2}{*}{Strategy} & 
\multicolumn{3}{c|}{Test-8} &
\multicolumn{3}{c}{Test-74} \\
\cline{2-7}
&Precision$\bm{\uparrow}$  &Recall$\bm{\uparrow}$   &F1$\bm{\uparrow}$ &Precision$\bm{\uparrow}$  &Recall$\bm{\uparrow}$   &F1$\bm{\uparrow}$ \\
\hline
Greedy                
&0.533     &0.584     &0.557
&0.535     &0.560     &0.544     \\
\rowcolor[HTML]{F2F2F2}
Self-corrective
&\textbf{0.676}     &\textbf{0.619}     &\textbf{0.631}
&\textbf{0.582}     &\textbf{0.670}    &\textbf{0.622}     \\
\bottomrule
\end{tabular}
}
\vspace{-4.0pt}
\end{table}

\noindent\textbf{Bi-directional Modeling.}
Notably, our self-paced training and self-correction mechanisms are specifically designed for bi-directional models and cannot be directly applied to autoregressive models.
For a fair comparison, we align our model architecture (e.g., input and hidden dimensions, etc.) with the autoregressive method TGT~\cite{argaw2024towards} and compare them in~\cref{tab:modeling_layernum}.
The bi-directional model learned by our method outperforms autoregression even using fewer layers. 
\begin{table}[h]
\vspace{-4.0pt}
\caption{Bi-directional model (ours) v.s. Autoregression (TGT).}
\label{tab:modeling_layernum}
\vspace{-4.0pt}
\centering
\tabcolsep=3pt
\small{
\begin{tabular}{c|ccccc}
\toprule
\multirow{2}{*}{Method} & 
\multicolumn{5}{c}{Test-ALL}\\
\cline{2-6}
&Precision$\bm{\uparrow}$  &Recall$\bm{\uparrow}$   &F1$\bm{\uparrow}$ &LD$\bm{\downarrow}$  &AA$\bm{\uparrow}$\\
\hline
TGT [1] 5-layer
& 0.1214     &0.1288     &0.1240
&96.08      &0.46\\
\hline
SSMP 4-layer 
&0.1492    &0.1734     &0.1601
&88.51     &0.66\\
\rowcolor[HTML]{F2F2F2}
SSMP 5-layer
&\textbf{0.1596}     &\textbf{0.1847}    &\textbf{0.1710}
&\textbf{87.93}     &\textbf{0.68}\\ 
SSMP 8-layer &0.1577   &0.1836     &0.1694 &88.09    &\textbf{0.68}\\
\bottomrule
\end{tabular}
}
\vspace{-4.0pt}
\end{table}

\noindent\textbf{Model architecture.}
As shown in~\cref{tab:modeling_layernum}, the 5-layer model achieves the best results. 
The 4-layer and 8-layer models suffer from under- and over-fitting, respectively. 
We adopt the 5-layer model, considering the current limited data scale.
In the future, we will expand the dataset for training a larger model.

\noindent\textbf{Movie Time Periods and Genres.}
\cref{tab:time_genre} shows that our method performs best for movies between 2010 and 2020, likely because most of the training data comes from this period.
In addition, our method achieves better performance on Crime, Thriller, and Comedy movies, while the performance on Adventure movies is relatively weaker. This may be attributed to the higher variability in visual content and narrative structure in Adventure films, as well as the potential genre imbalance in the training data.
\begin{table}[h]
\vspace{-4.0pt}
\caption{Results on movies across time periods and genres.}
\label{tab:time_genre}
\vspace{-4.0pt}
\centering
\tabcolsep=2pt
\small{
\begin{tabular}{@{}cc|ccccc@{}}
\toprule
\multirow{2}{*}{\begin{tabular}{c}
Time Periods \\
Genres
\end{tabular}} & \multirow{2}{*}{\#Movies} &
\multicolumn{5}{c}{Test-ALL}\\
\cline{3-7}
&
&Precision$\bm{\uparrow}$  &Recall$\bm{\uparrow}$   &F1$\bm{\uparrow}$ &LD$\bm{\downarrow}$  &AA$\bm{\uparrow}$\\
\hline
2000-2010 &25
&0.1726     &0.1971     &0.1838
&93.86      &0.68\\
2010-2020 &57
&\textbf{0.1920}    &\textbf{0.2220}     &\textbf{0.2056}
&\textbf{88.69}     &\textbf{0.69}\\
after 2020 &30
&0.1587    &0.1999     &0.1759
&93.29     &0.60\\
\hline
Thriller &43
&0.2031     &0.2259     &0.2132
&90.56      &\textbf{0.68}\\
Action &30
&0.1964     &0.2172     &0.2060
&93.25      &0.65\\
Comedy &30
&0.2104     &0.2400     &0.2240
&91.75      &0.62\\
Adventure &23
&0.1742     &0.2041     &0.1877
&83.56      &\textbf{0.68}\\
Crime &23
&\textbf{0.2311}     &\textbf{0.2543}     &\textbf{0.2419}
&\textbf{82.53}      &0.66\\
\bottomrule
\end{tabular}
}
\end{table}

\end{document}